\documentclass[letterpaper, 10 pt, conference]{ieeeconf}  
\IEEEoverridecommandlockouts                              
\usepackage{balance}
\usepackage{newtxtext}
\usepackage{amsmath} 
\usepackage{amssymb}  
\usepackage{mathrsfs}  

\usepackage[pdftex]{graphicx}
\graphicspath{{./images/}}
\DeclareGraphicsExtensions{.pdf,.jpeg,.png}

\usepackage{algorithm}
\usepackage{algpseudocode}

\makeatletter
\let\MYcaption\@makecaption
\makeatother

\usepackage[font=footnotesize]{subcaption}

\makeatletter
\let\@makecaption\MYcaption
\makeatother

\usepackage{color} 

\newcommand{\code}[1]{\texttt{#1}}

\usepackage{bbm}
\usepackage{xfrac}
\usepackage{siunitx}

\usepackage{cite}

\usepackage{letltxmacro}
\LetLtxMacro{\vec}{\vector}
\renewcommand{\vec}[1]{\mathbf{#1}}

\newcommand{\transpose}{^\top}
\newcommand{\func}[1]{\textit{#1}}

\newcommand{\prm}{PRM$^*$}
\newcommand{\rrt}{RRT$^*$}
\newcommand{\fmt}{FMT$^*$}

\DeclareMathOperator*{\argmin}{arg\,min}

\DeclareMathOperator{\atantwo}{atan2}

\usepackage{amsthm}
\newtheorem{problem}{Problem}

\theoremstyle{definition}

\newtheorem{remark}{Remark}

\title{\LARGE \bf
	Streamlines for Motion Planning in Underwater Currents
}
\author{K.~Y.~Cadmus~To$^1$,
        Ki~Myung~Brian~Lee$^1$,
        Chanyeol~Yoo$^1$,
        Stuart Anstee$^2$
        and Robert~Fitch$^1$
    \thanks{This research is supported by an Australian Government Research Training Program (RTP) Scholarship, Australia's Defence Science and Technology Group, Bureau of Meteorology, and the University of Technology Sydney.}
	\thanks{$^1$Authors are with the University of Technology Sydney, Ultimo, NSW 2006, Australia {\tt\footnotesize \{Cadmus.To,Brian.Lee\}@student.uts.edu.au} and {\tt\footnotesize \{Chanyeol.Yoo,Robert.Fitch\}@uts.edu.au}}
	\thanks{$^2$Author is with the Defence Science and Technology Group, Department of Defence, Australia {\tt\footnotesize stuart.anstee@dst.defence.gov.au}}
    }

\begin{document}
	\thispagestyle{empty}
	\pagestyle{empty}

	\maketitle

\begin{abstract}
Motion planning for underwater vehicles must consider the effect of ocean currents.
We present an efficient method to compute reachability and cost between sample points in sampling-based motion planning that supports long-range planning over hundreds of kilometres in complicated flows.
The idea is to search a reduced space of control inputs that consists of stream functions whose level sets, or streamlines, optimally connect two given points.
Such stream functions are generated by superimposing a control input onto the underlying current flow.
A streamline represents the resulting path that a vehicle would follow as it is carried along by the current given that control input.
We provide rigorous analysis that shows how our method avoids exhaustive search of the control space, and demonstrate simulated examples in complicated flows including a traversal along the east coast of Australia, using actual current predictions, between Sydney and Brisbane.
\end{abstract}

	\section{Introduction}

Underwater vehicles are important in many applications including environmental monitoring~\cite{Rudnick2004}, oil and gas exploration~\cite{Russell-Cargill2018}, and defence~\cite{Johannsson2010}. While the majority of robotic ocean sensors drift freely with the current~\cite{Argo2000}, the requirement to concentrate sensing in areas of high priority has led to increased interest in buoyancy-driven autonomous underwater gliders~\cite{Webb2001,Bachmayer2004}, propeller-driven autonomous underwater vehicles (AUVs)~\cite{Stokey2005}, and hydrids of the two~\cite{Claus2010}.
We are interested in improving the autonomous operation of underwater vehicles by introducing a new planning tool that gains computational efficiency by exploiting a concept in fluid dynamics called a \emph{stream function}.

The motion of underwater vehicles is heavily influenced by prevailing ocean currents. Optimal motion planning for underwater vehicles can be viewed as an instance of the long-standing \emph{Zermelo's Problem}~\cite{Zermelo_RefBook1931}, for which there is no known efficient analytical solution in general.
A numerical approach is to apply the well-known \emph{shooting method}, which can be used within a sampling-based planning framework to find edge connections and costs. We use such a method, for example, within \fmt{} to produce an asymptotically optimal minimum-energy planner for gliders in our recent work~\cite{lee2017energy}.
However, an open challenge is how to efficiently find edge connections in this approach. The shooting method involves forward integration along the edge, which must be repeated for each of a set of control values to solve the underlying two-point boundary value problem. This high computational cost limits the number of edges that can be evaluated in practice, which in turn limits the solution quality or geographical scale of problem instances that can be feasibly solved.

In this paper, we examine the use of stream functions within a sampling-based planner to find edge connections.
Level sets of the stream function, known as \emph{streamlines}, represent paths that a vehicle would follow as it is carried along by the ocean current in the absence of any control input. We propose control inputs that induce a new stream function that, when superimposed on the underlying current, acts as a kind of local roadmap. We show how to efficiently search the space of control actions so that a streamline path between two points is found, if one exists. Forward integration of control inputs is still required, but our method avoids exhaustive search in control space and hence gains a major computational advantage by performing far fewer integrations than in a typical shooting method implementation.

We integrate our streamline method with \prm{} and provide rigorous computational analysis. The planning problem is formulated in the horizontal plane, and we demonstrate paths through flow fields with multiple gyres; some over hundreds of kilometres. The main significance of our result is that the streamline method can be used with a variety of asymptotically optimal planning algorithms, such as \rrt{}, \fmt{}, and other variants, to efficiently find high quality paths with sparse control inputs.
We focus here on time-optimal paths, but other objectives such as minimum energy can be accommodated by modifying the objective function of our streamline method.

	\section{Related work}

Mission plans for underwater vehicles must allow for the oceanic flows through which they travel, and predictions of three-dimensional~(3D) oceanic flow fields are freely available from multiple sources~\cite{Oke2005, Oke2013, Shchepetkin2005}. A method that estimates flows from observations is given by~\cite{Chang2017}.

Although optimal planning in flow fields is generally well-studied, existing methods exhibit critical limitations.
Work most closely related to ours elegantly uses level-set methods to find time-optimal~\cite{Lolla2014a} or energy-optimal~\cite{subramani2016energy} paths by explicitly solving for the reachable set as a level set of a scalar function. A numerical solution of the partial differential equation involved, however, can be computationally prohibitive in robotics applications.
A variety of graph-based methods have been proposed, where the workspace is uniformly~\cite{Kularatne2012} or adaptively~\cite{Kularatne2018c} discretised. This class of algorithms is resolution-complete, and is subject to well-known performance trade-offs and computational drawbacks~\cite{LaValle2006}.
Related work that uses sampling-based planning includes application of the RRT algorithm~\cite{Ko2014}, where expansions are biased to follow current flow.
Our previous work presents an FMT*-based energy-optimal algorithm in 3D where edges represent glider trim states computed using the shooting method~\cite{lee2017energy}.

Here, we propose a principled method to reduce the size of the control space when computing reachability between sample points. The use of numerical computation is drastically reduced compared to fully computing reachable sets, and reachability computation is performed relatively infrequently compared to graph-based methods with dense resolution. Our method thus supports long-range navigation as we demonstrate in this paper. 


\section{Background}

\subsection{Vehicle model}
Suppose we represent the motion of a vehicle $G$ by a continuous-time transition model
\begin{equation} \label{eqn:contSystem}
	\dot{\vec{x}} = F_v\left(\vec{x}\right) + \vec{v}_g
	,
\end{equation}
%
%
where $\vec{x} = \left[x,y\right]\transpose$ is the position of the glider in two-dimensional~(2D) space, $F_v(\mathbf{x}) = [u_f, v_f]\transpose$ is the time-independent flow vector and~$\vec{v}_g = \left[u_g,v_g\right]\transpose$ is the glider's velocity \emph{relative} to the flow, which is bounded by maximum speed~$V_{max}$ such that
\begin{equation} \label{eqn:filledControlCircle}
	|\mathbf{v}_g| \leq V_{max}
	.
\end{equation}

The continuous system~\eqref{eqn:contSystem} can be discretised with a small time step~$\Delta t$ as
\begin{equation}  \label{eqn:discreteSystem}
	\vec{x}_{k+1} = \vec{x}_{k} + \left( F_v\left(\vec{x}_{k}\right) + \vec{v}_{g,k} \right) \Delta t
	.
\end{equation}
The discrete system is controlled by action~$a_k = \mathbf{v}_{g,k}$ at $k$-th time step. In a short form, we have~$f_{\mathrm{x}}(\mathbf{x}_k, a_k)$.
The sequence of vehicle controls~$\sigma$ is denoted as
\begin{equation}
	\sigma = a_0 a_1 \cdots a_{K-1}
    ,
\end{equation}
where~$K$ is the discrete time horizon. 
We denote~$f_{\mathrm{x}}(\mathbf{x}, \sigma)$ as the resultant state after executing the control sequence~$\sigma$ from~$\mathbf{x}$.
Note that a control action~${a_k \in \mathcal{A}}$ is constrained by the upper limit on the reference velocity~$\mathbf{v}_g$
\begin{equation} \label{eqn:control_set_simple}
	\mathcal{A} = \{ \mathbf{v} \in \mathbb{R}^2 \mid |\mathbf{v}| \leq V_{max} \}
    .
\end{equation}

The cost of transiting from state~$\mathbf{x}$ with control~$a$ is denoted as~$f_c(\mathbf{x}, a)$. The cost for control sequence~$\sigma$ from state~$\mathbf{x}$ is denoted as~$f_c(\mathbf{x}, \sigma)$.



\subsection{Stream functions}



We consider a time-invariant, incompressible flow field~\cite{Pritchard2011}
\begin{equation}\label{eqn:incompressibility}
	\nabla \cdot F_v = 0
    ,
\end{equation}
where~$\nabla \cdot$ is the divergence operator.
With this condition, which requires that density does not change with flow, a 2D flow field can be represented using the \emph{stream function}~${\psi: \mathbb{R}^2 \times \mathbb{R}^2 \rightarrow \mathbb{R}}$, which describes the flux that passes through a curve connecting two points in 2D space~\cite{Batchelor1967}.
The value of the stream function~$\psi_{PQ}$, or \emph{stream value} between two points~$P$ and~$Q$ is
\begin{equation} \label{eqn:defStreamfunction}
	\psi\left(\vec{x}_P,\vec{x}_Q\right) = \int_{\vec{x}_P}^{\vec{x}_Q} \left( u_f\left(\vec{x}\right) dy - v_f\left(\vec{x}\right) dx \right)
	.
\end{equation}

\begin{figure}[t]
	\centering
	\includegraphics[width=\columnwidth]{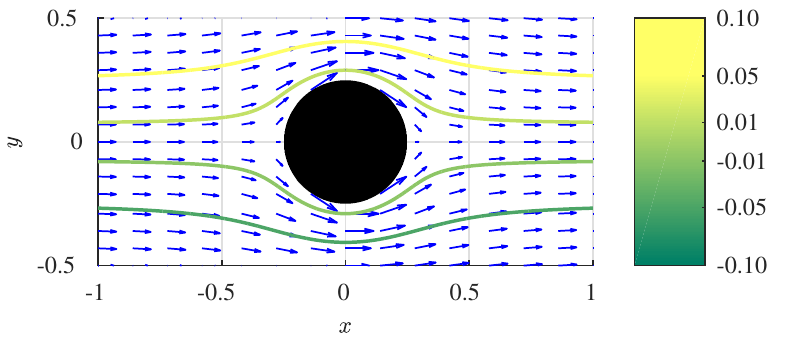}
    \caption{Incompressible flow visualised by streamlines. The stream values of the streamlines refer to the flux (rate of flow) of the fluid flowing between each streamline and the arbitrary reference point $\left(-1,0\right)$.}
    \label{fig:streamlines}
\end{figure}

If the flow field is incompressible (i.e. satisfies \eqref{eqn:incompressibility}), then the line integral \eqref{eqn:defStreamfunction} is well-defined in the sense that the value is the same for any curve connecting $\mathbf{x}_{P}$ and $\mathbf{x}_{Q}$. 
A continuous set of points that have the same stream value relative to some reference point~$P$ is known as a~\emph{streamline}. The term is apt, because the stream value of an idle vehicle (i.e. $\mathbf{v}_{g} = \mathbf{0}$) advected by the flow field remains constant. In other words, we have:  
\begin{equation}\label{eqn:constantStreamValue}
\begin{aligned}
	\frac{d}{dt} \psi( \mathbf{x}_{P}, \mathbf{x}_{t} ) &= -v_{f} u_{f} + u_{f} v_{f} = 0
    .
\end{aligned}
\end{equation}
In Fig.~\ref{fig:streamlines}, four arbitrary streamlines with different stream values illustrate the flow of fluid around a circular obstacle. 
Disjointed streamlines with the same stream value are referred to as being \emph{distinct}.

\begin{remark} [Additive property of stream function] \label{remark:additive}
A useful property of incompressible flow fields is that they can be represented by the sum of their stream functions: given two incompressible flow fields~$F_A$ and~$F_B$ with corresponding stream functions~$\psi_A$ and~$\psi_B$, the superimposed flow field~$F_{A+B}$ is
    \begin{equation}
        \func{F}_{A+B} = \func{F}_A + \func{F}_B
        ,
    \end{equation}
    and the corresponding stream value~$\psi_{A+B}$ is:
    \begin{equation}
        \psi_{A+B} = \psi_{A} + \psi_{B}
        .
    \end{equation}
\end{remark}
This property becomes vital in Sec.~\ref{sec:bound} to gain deeper insight to the vehicle's net trajectory by considering the net stream function after treating the vehicle's relative velocity as a flow field.

\begin{figure*}[t]
	\centering
	\begin{subfigure}[b]{0.24\textwidth}
		\centering
    	\includegraphics[height=1\columnwidth]{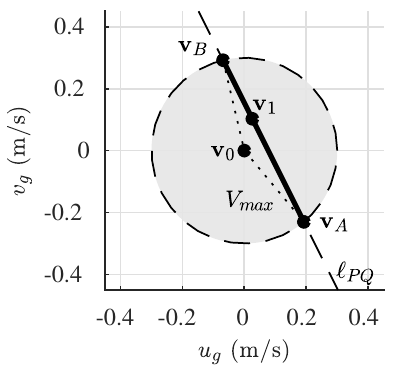}
		\caption{Constrained control samples}
		\label{fig:controlIntersection}
	\end{subfigure}
	\begin{subfigure}[b]{0.24\textwidth}
		\centering
		\includegraphics[height=1\columnwidth]{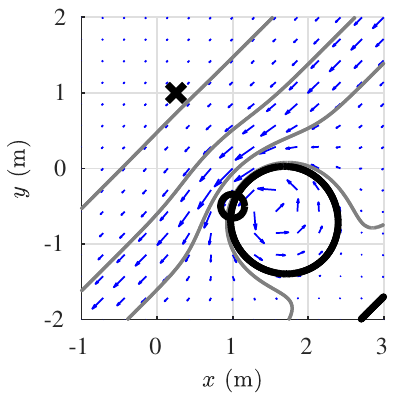}
		\caption{$\vec{v}_g = \vec{v}_0$}
		\label{fig:v0}
	\end{subfigure}
	\begin{subfigure}[b]{0.24\textwidth}
		\centering
		\includegraphics[height=1\columnwidth]{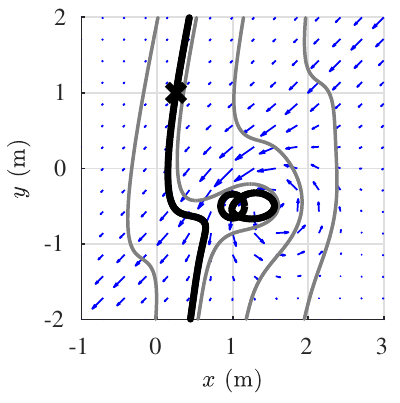}
		\caption{$\vec{v}_g = \vec{v}_1$}
		\label{fig:v1}
	\end{subfigure}
	\begin{subfigure}[b]{0.24\textwidth}
		\centering
		\includegraphics[height=1\columnwidth]{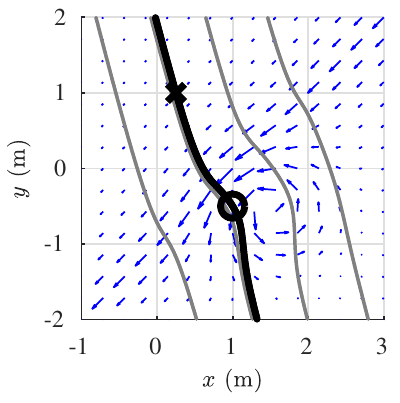}
		\caption{$\vec{v}_g =\vec{v}_B$}
		\label{fig:v2}
	\end{subfigure}
	\caption{Streamlines resulting from superimposing stream functions of the environmental flow and the vehicle's relative velocity.}
	\label{fig:combinedStreamlines}
\end{figure*}

\section{Problem statement}

We consider a path planning problem for vehicle~$G$. We seek an optimal control sequence~$\sigma^* \in \mathcal{A}^K$ over a cost function~$f_c$ from initial position~$\vec{x}_{init}$ to goal position~$\vec{x}_{goal}$ in the presence of an incompressible flow field~$F_v$.

%
%

Motivated by the need to control underwater vehicles with limited energy capacity and connectivity~\cite{dhanak2016springer}, we are interested in a scenario which minimises changes in control inputs to reduce energy expenditure that would otherwise be spent on computation. With this constraint, the vehicle executes a sequences of \emph{persistent} controls~$\omega$
\begin{equation}
	\omega = (a_0, \tau_0), \cdots, (a_{K-1}, \tau_{K-1})
    ,
\end{equation}
where a \emph{persistent control}~$\omega_i$ consists of a control action~${a_i \in \mathcal{A}}$ with a duration~$\tau_i$. The key difference between~$\sigma$ and~$\omega$ is that the actions resulting from~$\sigma$ are separated by constant sampling time~$\Delta t$, whereas those from~$\omega$ coincide with decision points; for example, when a vehicle surfaces for a position fix.
We slightly abuse the notation for cost such that~$f_c(\mathbf{x}_k, \omega_k)$ denotes the cost of executing persistent control~$\omega_k$ starting from~$\mathbf{x}_k$ and~$f_c(\mathbf{x}_{init}, \omega)$ denotes the cost of executing the entire sequence~$\omega$.
Similarly, we denote by~$f_{\mathrm{x}}(\mathbf{x}_{init}, \omega)$ the resultant state after executing~$\omega$.
We assume that any control duration~$\tau_i$ is of sufficient length so that the control transition time from~$a_i$ and~$a_{i+1}$ is negligible.

The path planning problem with a persistent control sequence is formally defined as follows:
\begin{problem} [Energy-aware optimal path planning in a flow field]
	Given a vehicle~$G$, a set of controls~$\mathcal{A}$, initial position~$\mathbf{x}_{init}$, goal position~$\mathbf{x}_{goal}$, and time-invariant incompressible flow field~$F_v$, find an energy-aware disjoint control sequence~$\omega^*$ that minimises the overall cost from~$\mathbf{x}_{init}$ to~$\mathbf{x}_{goal}$ such that
	\begin{equation}
        \omega^* = \argmin_{\omega} \sum_{k = 0}^{K-1} f_c \left( \mathbf{x}_{k}, \omega_k \right)
        ,
    \end{equation}
    where~$\mathbf{x}_{k+1} = f_{\mathrm{x}}(\mathbf{x}_k, \omega_k)$, $\mathbf{x}_{0} = \mathbf{x}_{init}$, and $\mathbf{x}_{K} = \mathbf{x}_{goal}$.
\end{problem}

This problem consists of two sub-problems: 1) find a persistent control connecting two arbitrary points in the flow field (if it exists), and 2) find a sequence that minimises the overall cost. Without loss of generality, we consider time cost in this paper.

The key aspect of this formulation is that it fits naturally with sampling-based planning methods such as \prm{}~where samples are separated in space and edge connections are made between pairs of sample points. We show that our proposed method reduces a complexity bottleneck inherent in making the edge connections in a flow field by significantly reducing the control space.


\section{Streamline-based control search}

In this section, we present a novel method for efficiently finding a persistent control between two arbitrary points in the presence of an incompressible time-invariant flow field. 
We significantly reduce the control space by exploiting the idea of stream functions.

\subsection{Finding control bound using stream function} \label{sec:bound}
The vehicle~$G$ is controlled by adapting its relative velocity~$\mathbf{v}_g$. The relative velocity can be viewed as an additional flow acting on an idle vehicle. We denote such flow~$\func{F}_g = \vec{v}_g$ and we call~$F_g$~the \emph{flow due to control}.
Formally, the vehicle is in an equivalent reference frame with a flow field~$(F_v + F_g)$ and control~$|\bar{\mathbf{v}}_g|=0$.
From~\eqref{eqn:defStreamfunction}, for \emph{constant} relative velocity $\vec{v}_g$, the stream value for~$F_g$ between two points~$P$ and~$Q$ is simply
\begin{equation}
	\psi_{g}\left(\vec{x}_P,\vec{x}_Q\right) = u_g\Delta y - v_g\Delta x,
\end{equation}
where $\Delta x = x_Q-x_P$ and $\Delta y = y_Q-y_P$.

By Remark~\ref{remark:additive}, the superimposed stream value for external flow~$F_v$ and the flow due to vehicle motion~$F_g$ is
\begin{equation}
	\psi_{PQ} \equiv \psi\left(\vec{x}_P,\vec{x}_Q\right) = \psi_{v}\left(\vec{x}_P,\vec{x}_Q\right) + u_g\Delta y - v_g\Delta x
    .
\end{equation}

The aim is to find a control~$\vec{v}_g$ such that two points~$P$ and~$Q$ are on the same streamline in the superimposed flows, that is, $\psi_{PQ} = 0$ or
\begin{equation} 	\label{eqn:isoline}
	\psi_{v}\left(\vec{x}_P,\vec{x}_Q\right) + u_g\Delta y - v_g\Delta x = 0
	.
\end{equation}
Note that all points on the same streamline have the same stream value, but the converse need not apply, since points with the same stream value may reside on distinct streamlines as shown in Fig.~\ref{fig:v1}.

In~(\ref{eqn:isoline}), we defined a line in the control space over~$u_g$ and~$v_g$ that connects points~$P$ and $Q$. We denote it by~$\ell_{PQ}$.
The streamline-based set of \emph{feasible} controls for manoeuvring from~$P$ to~$Q$ is given as
\begin{equation} \label{eqn:control_set}
	\mathcal{A}_{PQ} = \{ \vec{v} \in \ell_{PQ} \mid \left|\vec{v}\right| \leq V_{max} \}
    .
\end{equation}
Importantly, with streamline-based control, we need only search a one-dimensional (1D) control space; in our previous approach~\cite{lee2017energy}, which was based on~\eqref{eqn:control_set_simple}, the space was 2D. We discuss the reduction in complexity in Sec.~\ref{sec:analysis}.

Since the set of controls satisfying~$|\mathbf{v}_g| \leq V_{max}$ is convex and~$\ell_{PQ}$ is linear in control space, three sets of controls are possible depending on the number of intersections between the two conditions: 1) the empty set if there is no intersection, 2) a set with one element if there is only one intersection, and 3) a set of infinite solutions between two intersections. Defining the intersection points in control space as~\emph{endpoints}~$\vec{v}_A$ and~$\vec{v}_B$, we find
\begin{equation}
	\begin{aligned}
		\vec{v}_A &= V_{max}\left[\cos{\theta_A},\sin{\theta_A}\right]\transpose \\
		\vec{v}_B &= V_{max}\left[\cos{\theta_B},\sin{\theta_B}\right]\transpose
	\end{aligned}
	,
\end{equation}
where
\begin{equation}
	\begin{aligned}
		\theta_A &= \delta + \frac{\pi}{2} + \arccos{\left(\kappa\right)} \\
		\theta_B &= \delta + \frac{\pi}{2} - \arccos{\left(\kappa\right)} \\
		\delta &= \atantwo \left(\Delta y, \Delta x\right) \\
		\kappa &= \frac{\psi_{v}\left(\vec{x}_P,\vec{x}_Q\right)}{V_{max}\left|\left|\vec{x}_Q- \vec{x}_P\right|\right|_2}
	\end{aligned}
	.
	\label{eqn:streamIntervalEndpoints}
\end{equation}
Intuitively, the set of feasible controls lies along the straight line between the endpoints.

In Fig.~\ref{fig:combinedStreamlines}, we illustrate the streamlines for different controls~$\vec{v}_g \in \mathcal{A}_{PQ}$. The control space shown in Fig.~\ref{fig:controlIntersection} includes the maximum speed constraint~$|\vec{v}_g| \leq V_{max}$, the control line~$\ell_{PQ}$, and the endpoints~$\vec{v}_A$ and~$\vec{v}_B$ of its feasible subset. 
In Fig.~\ref{fig:v0}, the control~$\vec{v}_g = \vec{v}_0$ is not on the control line~$\ell_{PQ}$, so the points~$P$ and~$Q$ have different stream values and one is not reachable from the other.
In Fig.~\ref{fig:v1}, the control is on the control line and points~$P$ and~$Q$ have the same stream value but are on distinct streamlines.
In Fig.~\ref{fig:v2}, the control is on the control line and point~$Q$ is downstream from~$P$, and is therefore reachable.

\subsection{Finding critical points in a control set} \label{sec:critical}

We have shown that endpoints~$\vec{v}_A$ and~$\vec{v}_B$ define a set of feasible controls~$\mathcal{A}_{PQ}$ that guarantees points~$P$ and~$Q$ have the same stream values. However, as Fig.~\ref{fig:v1} shows, the endpoint-based method does not guarantee that they are on the same streamline.
Note that the points are guaranteed to be unreachable if they have different stream values.


It turns out that distinct streamlines originate at saddle points in the flow.
This can be shown through~\emph{Morse theory}~\cite{Smith2013}, the study of critical points of a smooth function. 
Given two points~$P$ and~$Q$ in state space and flow field~$F$, a point~$\mathbf{c}$ is a saddle point iff~\cite{Smith2013}
\begin{equation} \begin{aligned}
	F(\vec{c}) = 0~\text{and}~\lambda_1 \lambda_2 < 0
    ,
    \label{eqn:stop_cond}
\end{aligned} \end{equation}
where~$\lambda_1$ and $\lambda_2$ are the eigenvalues of the Hessian matrix~$\Delta \psi(\vec{c})$. 
Intuitively, a point is a saddle point iff the flow at the point is idle but is at neither a maximum nor a minimum. 
A result from Morse theory is that level sets, such as the streamlines we consider in this paper, are \emph{connected} if there are no saddle points \cite{Smith2013}; that is, the sampling strategy~\eqref{eqn:streamIntervalEndpoints} then guarantees that the vehicle will travel from $P$ to $Q$. 
A simple way to check for the existence of a saddle point is to check the magnitude of the net velocity. If it is close to zero and the determinant of the Hessian of the stream function is negative, then a saddle point exists near the trajectory. 
In physical terms, the saddle points imply that the vehicle comes to a `stall'. This is clearly undesirable in either time- or energy- optimal case. 

The condition~\eqref{eqn:stop_cond} thus implies that two points~$P$ and~$Q$ lie on the same streamline if the net velocity of the vehicle with respect to absolute reference frame (i.e., $F_v + F_g$) is sufficiently large over the streamline (i.e., trajectory) from~$P$ to~$Q$.
Importantly, this serves as a stopping condition that further reduces the time complexity.

\subsection{Control space sampling} \label{sec:control_sampling}

The reduced control space and the saddle point-based stopping condition~\eqref{eqn:stop_cond} allow us to find the control~$a^* \in \mathcal{A}$ that minimises the objective function.
Two steps are required to find the optimal control between two states: control sampling and forward integration. 

Given a control line~$\ell_{PQ}$, we linearly sample $C$~controls between and inclusive of endpoints~$\vec{v}_A$ and~$\vec{v}_B$. For each control, we find a continuous trajectory over state space by forward integrating the state of the vehicle from point~$P$ to~$Q$ based on~\eqref{eqn:discreteSystem}. 
We continue the integration until any of the following conditions holds true: 1) the vehicle is near~$Q$ (i.e., destination reached), 2) the integration has exceeded a specified maximum time horizon~$H$, or 3) the vehicle has reached a saddle point (i.e., an infeasible control).
Once we complete the enumeration over a set of controls for a given state sample pair, we find the control that minimises the travel time among the set of controls that reached the destination.

\begin{figure*}[t]
    \centering
    \begin{subfigure}[b]{0.49\columnwidth}
    	\centering
        \includegraphics[width=\textwidth]{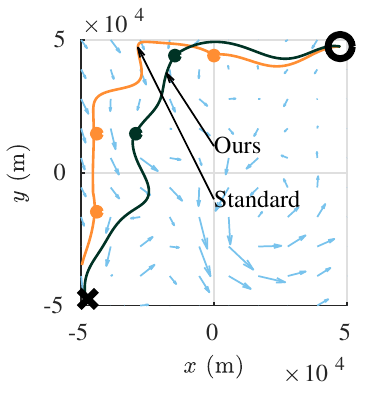}
        \caption{Std: 8.1~days, Ours: 3.9~days}
        \label{subfig:compositeTR}
    \end{subfigure}
    \begin{subfigure}[b]{0.49\columnwidth}
    	\centering
        \includegraphics[width=\textwidth]{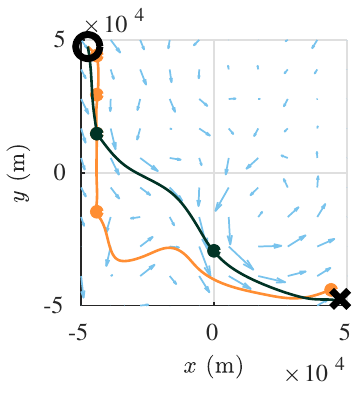}
        \caption{Std: 3.5~days, Ours: 2.2~days}
        \label{subfig:compositeTL}
    \end{subfigure}
    \begin{subfigure}[b]{0.49\columnwidth}
    	\centering
        \includegraphics[width=\textwidth]{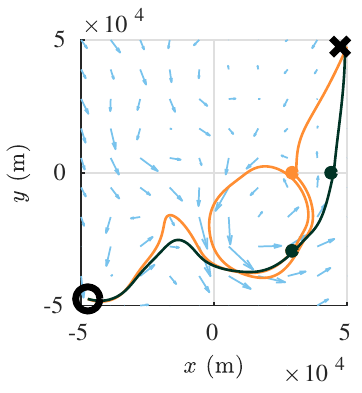}
        \caption{Std: 7.9~days, Ours: 3.4~days}
        \label{subfig:compositeBL}
    \end{subfigure}
    \begin{subfigure}[b]{0.49\columnwidth}
    	\centering
        \includegraphics[width=\textwidth]{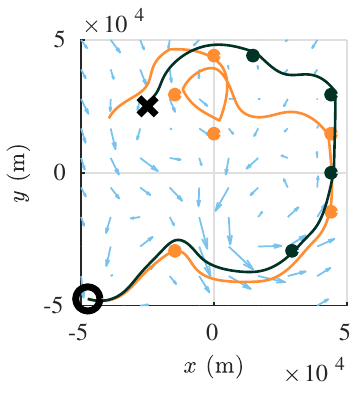}
        \caption{Std: 10.4~days, Ours: 5.4~days}
        \label{subfig:compositeBLTL}
    \end{subfigure}
    \caption{Minimum time trajectories using the \prm{} with streamline-based control search (dark green) and standard control search (orange) for a vehicle (circle) to reach its goal (cross). Intermediate waypoints (dots) are used to achieve this despite the limited speed of vehicle. $N=49$, $C=19$}
    \label{fig:composite}
\end{figure*}
\begin{figure*}[t]
    \centering
    \begin{subfigure}[b]{0.48\textwidth}
    	\centering
        \includegraphics[width=\textwidth]{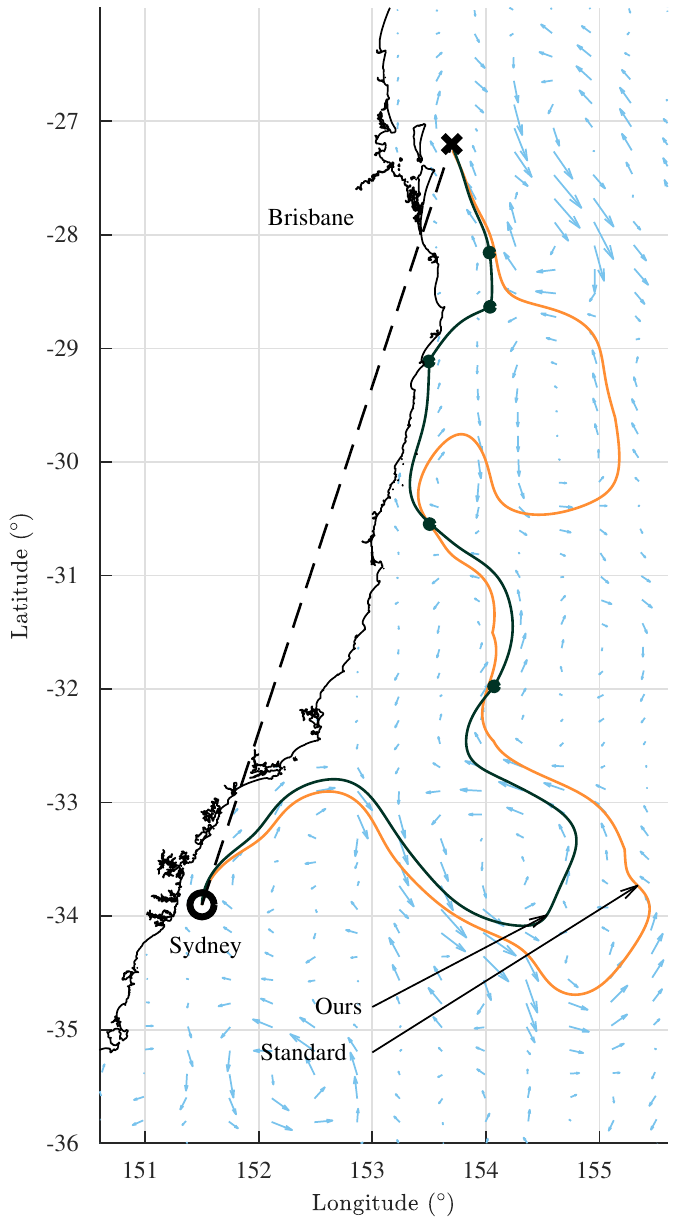}
        \caption{From Sydney to Brisbane}
        \label{subfig:syd_bne}
    \end{subfigure}%
    \begin{subfigure}[b]{0.48\textwidth}
    	\centering
        \includegraphics[width=\textwidth]{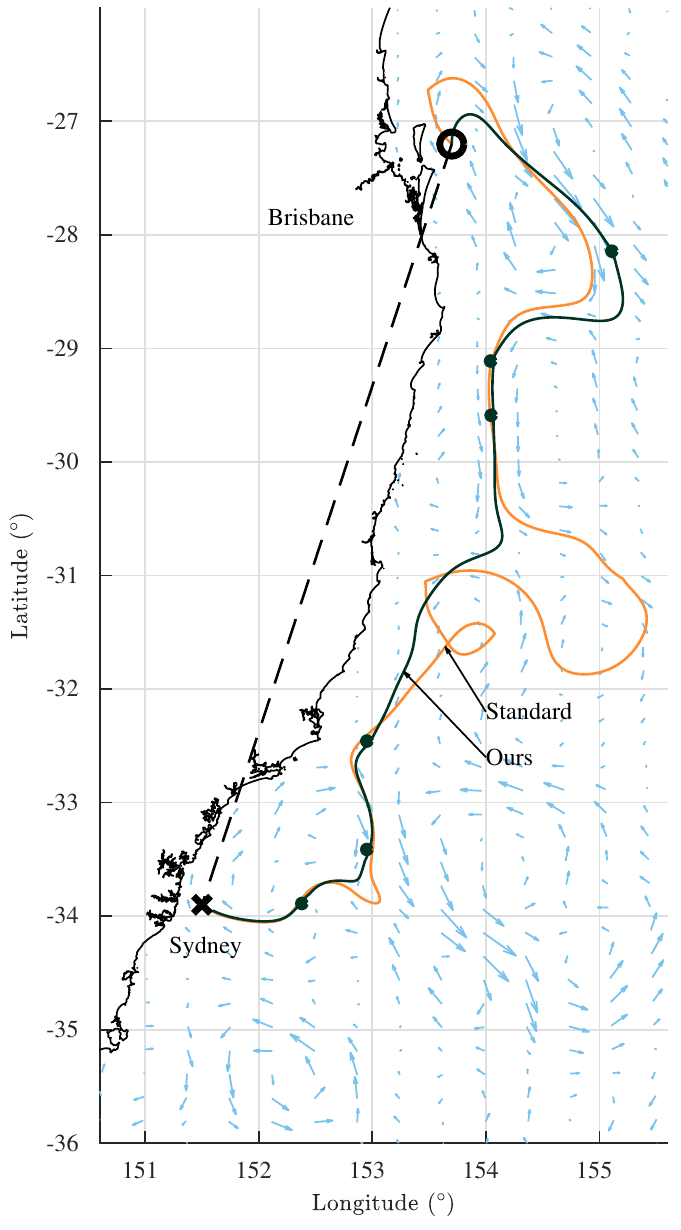}
        \caption{From Brisbane to Sydney}
        \label{subfig:bne_syd}
    \end{subfigure}
    \caption{\prm{} on ocean currents based on real dataset, starting at the circle and finishing at the cross. Waypoints for the streamline-based method (in dark green) are marked with dots, the waypoints for the standard shooting method (in orange) are similarly spaced.}
    \label{fig:nciResults}
\end{figure*}

\subsection{Streamline-based motion planning}

We present a high-level analysis of a~\emph{probabilistic roadmap$^*$} (\prm{}) approach to path planning in a flow field, in which the evaluation of edge costs for pairs of points in state space is a computational bottleneck.
Other sampling-based motion planners such as~\rrt{} and \fmt{} would show similar reductions in edge cost complexity.

The \prm{} algorithm randomly selects pairs of samples in state space. For each directed pair of state samples~$P$ and~$Q$, we find the edge time cost~$\code{Cost}(\vec{x}_P, \vec{x}_Q)$ using the streamline-based method.
If there exists no solution for a pair, the samples are not connected. Once we find all the necessary edge costs, we use Dijkstra's algorithm over the graph to find an optimal overall path from~$\vec{x}_{init}$ to~$\vec{x}_{goal}$.



\section{Analysis} \label{sec:analysis}


Given the number of state samples~$N$, the number of control samples for a state sample pair~$C$ and the time horizon for forward integration~$H$,
the worst-case time complexity of the framework is~$\mathcal{O}\left( N^2 \cdot C \cdot H \right)$. Although the complexity for the standard shooting method and the proposed method are similar, there are fundamental differences that make our framework significantly more efficient. 

In our framework, controls sampled from a 1D line segment are guaranteed to have the same stream values. In contrast, the shooting method samples controls from a 2D control space bounded by the maximum speed. Because a state is guaranteed unreachable from another if the states do not have the same stream value, most of the control samples evaluated using the standard shooting are invalid. 
As a result, significantly fewer control samples fall on the control line~$\ell_{PQ}$ leading to numerous unnecessary forward integrations, substantially increasing the computation time.

An interesting empirical observation is that the time-optimal control for a given state pair seems to be the control with the maximum speed. In other words, the optimal control is at one of the endpoints~$\vec{v}_A$ or~$\vec{v}_B$. 
If the special case is correct, then the overall time complexity becomes~$\mathcal{O}\left( N^2 \cdot H \right)$. 

\section{Case studies}

In this section, we demonstrate our implementation of the streamline-based motion planning method with cases.
The first employs a simulated environment where we compare our framework against the standard to significant difference in performance.
We then use our framework to find time-optimal solution using a real ocean dataset provided by the Australian Bureau of Meteorology~(BoM). We show that our method works efficiently over a large scale with challenging ocean currents.
For both cases, we consider a vehicle with maximum speed of~$0.3~\si{ms^{-1}}$, time horizon~$H$ of~$2000$ steps, and step size of 750~\si{s}.

\subsection{Simulated environment}

In Fig.~\ref{fig:composite}, we have four pairs of starting and goal positions (i.e. from circle to cross) in a simulated environment where the maximum flow magnitude is $1$~\si{ms^{-1}}, which is well beyond the vehicle's maximum speed. We compare our streamline-based method (in orange) against the standard shooting method (in dark green) with sampling parameters~${N=49}$ and~${C=19}$. 

For all scenarios, our method found significantly better solutions than the standard shooting method.
This is because we consider a more focussed set of samples in control space that increases the chance of finding an optimal solution.
In general, the streamline-based solutions exploit the currents to travel faster, whereas the standard shooting-based solutions are less effective in doing so.
As discussed in Sec.~\ref{sec:analysis}, we found that the time-optimal controls lie at the endpoints. If this endpoint hypothesis is true, the computation time could be reduced by eight times in this case.



\subsection{Eastern Australian Currents}

We demonstrate the use of the streamline-based framework to find time-optimal paths between Sydney and Brisbane. It is important to note that the region is very challenging to operate a vehicle with relatively slow speed; there exists strong southward currents and a number of eddies where the vehicle simply cannot pass or would otherwise get trapped.

The dataset provided by BoM is generated by an ocean model using a series of satellite images measuring the ocean heights. In this case study, we use a dataset that estimates the currents on 5th September 2018. From the dataset, we numerically computed the stream values using~\eqref{eqn:defStreamfunction} by assuming that the ocean flow is incompressible. Note that the dataset includes ocean flows more than 7 times faster than the vehicle's maximum speed.

The resulting paths between Sydney and Brisbane are shown in Fig.~\ref{fig:nciResults}. We also compare our streamline-based method (in orange) against the standard shooting-based method (in dark green). For both methods, we use the sampling parameters~$N=210$ and~$C=19$. 

For the path from Sydney to Brisbane in Fig.~\ref{subfig:bne_syd}, both methods seem to take advantage of the currents (bottom right) before moving towards Brisbane. Although the travelled distance is similar, the streamline-based method took 17~days to reach Brisbane whereas the standard shooting-based method took 22.8~days. The vehicle took almost the same time (17.6 and 29.4~days, respectively) to travel back from Brisbane to Sydney shown in Fig.~\ref{subfig:bne_syd}.
For the purpose of comparing the results, the benchmark time it would take to travel along a straight line between Sydney and Brisbane in still water is 29.8~days. Our method clearly took a more efficient path, whereas the standard shooting-based method performed as poorly as this benchmark.


An important aspect of the proposed framework is that the control stays the same between two adjacent state samples (i.e. waypoints). Intuitively, once we set the control from the starting state, we let the vehicle go until it reaches the destination in the absence of active control.
This aspect is important, especially for underwater platforms where the energy capacity is limited and the energy consumption is related to changes in control~\cite{lee2017energy}. From Sydney to Brisbane, the vehicle changes its control only 5 times over 17 days. Graph-based methods, for example, would potentially apply a new control input at every time step.

\section{Conclusion and Future Work}
We have presented an algorithm to efficiently find a stream function induced by a control input superimposed on a flow field, such that two given points are connected by a streamline. We showed how this method can be integrated with a sampling-based algorithm to plan long-range paths for underwater vehicles in real-world ocean currents.

One limitation of our approach so far is that it is restricted to 2D planning. It is useful for gliders, for example, because typical onboard software accepts 2D waypoints as input and generates a depth profile automatically. However, stream functions are also defined in three dimensions, and it would be interesting to extend our approach in this way.

We have designed our method to be easily integrated with other sampling-based algorithms. Implementing FMT* or BIT*, for example, would be interesting avenues to pursue, in addition to performing field experiments with gliders and other types of AUVs. 
Beyond motion planning, the proposed method can be used in task planning for vehicles in flow fields~\cite{yoo2016online,yoo2012probabilistic}, which would benefit from the reduced complexity.

\balance
\bibliographystyle{IEEEtran}
\bibliography{IEEEabrv,ref,cadmus_paper,generic}

\end{document}